%% file: main.tex
\documentclass[11pt,a4paper]{article}
\usepackage[a4paper,left=22mm,right=22mm,top=27mm,bottom=23mm,headheight=14pt,headsep=9mm]{geometry}
\usepackage[T1]{fontenc}
\usepackage{XCharter}
\usepackage{microtype}
\usepackage[round,authoryear]{natbib}
\usepackage{titlesec,fancyhdr,caption}
\usepackage{floatrow,flafter}
\usepackage{enumitem}
\setlist{nosep,leftmargin=1.8em,topsep=4pt}
\titleformat{\section}{\large\bfseries}{\thesection.}{0.55em}{}
\titleformat{\subsection}{\normalsize\bfseries}{\thesubsection.}{0.55em}{}
\titleformat{\paragraph}[runin]{\normalsize\bfseries}{}{0pt}{}[]
\titlespacing*{\section}{0pt}{15pt plus 3pt minus 2pt}{7pt}
\titlespacing*{\subsection}{0pt}{11pt plus 2pt minus 1pt}{5pt}
\titlespacing*{\paragraph}{0pt}{5pt}{0.5em}
\renewcommand{\headrulewidth}{0.3pt}
\DeclareCaptionLabelSeparator{bar}{\enspace|\enspace}
\usepackage{graphicx,amsmath,amssymb,booktabs,tabularx,longtable,hyperref}
\usepackage{listings,xcolor}
\usepackage{tikz}
\usetikzlibrary{arrows.meta}
\definecolor{promptBackground}{HTML}{F1F6FE}
\definecolor{promptAccent}{HTML}{1D4EB0}
\title{SciMARS: Multi-Agent Recursive Self-Improvement for Closed-Loop Scientific Discovery}
\author{%
  {\normalsize\textbf{Weimin Wu}$^{\dagger}$, \quad \textbf{Han Liu}$^{\dagger,\S}$}\\[4pt]
  $^\dagger\;$Center for Foundation Models and Generative AI, Northwestern University\\
  Department of Computer Science, Northwestern University\\
  $^\S\;$Department of Statistics and Data Science, Northwestern University\\
  \texttt{\href{mailto:wwm@u.northwestern.edu}{\textcolor{red!70!black}{wwm@u.northwestern.edu}}},
  \texttt{\href{mailto:hanliu@northwestern.edu}{\textcolor{red!70!black}{hanliu@northwestern.edu}}}
}
\date{2026-11-08}
\hypersetup{colorlinks=true,linkcolor=blue!55!black,citecolor=blue!55!black,urlcolor=blue!60!black,pdftitle={SciMARS: Multi-Agent Recursive Self-Improvement for Closed-Loop Scientific Discovery}}
\makeatletter
\renewcommand{\maketitle}{\thispagestyle{plain}\noindent\makebox[0pt][l]{\makebox[\linewidth][r]{\raisebox{7pt}[0pt][0pt]{\footnotesize\itshape\@date}}}\rule{\linewidth}{0.3pt}\par\vspace{10pt}{\raggedright\fontsize{20}{24}\selectfont\bfseries\@title\par}\vspace{9pt}{\small\@author\par}\vspace{14pt}}
\makeatother
\renewenvironment{abstract}{\begingroup\normalsize\bfseries\setlength{\parskip}{0pt}}{\par\endgroup\vspace{10pt}}
\begin{document}
\maketitle
\begin{abstract}
We propose SciMARS, a framework for recursive task-local adaptation of multi-agent harnesses in closed-loop scientific discovery. Multi-agent systems coordinate analysis, experimental design, and verification as scientific evidence evolves. However, two problems limit their effectiveness: (i) without shared memory, agents lose observations and repeat work; (ii) fixed role procedures, handoffs, and memory policies perpetuate mistakes despite retained evidence. We address these problems with a shared evidence graph and separate component editors. SciMARS adapts roles and coordination; SciMARS + Memory adds a third editor for how agents record, retrieve, and receive evidence. Candidate probes guide composition and selection of the next harness. We evaluate scientific discovery through experimental-data replay and a digital twin. SciMARS + Memory resolves 12/18 GPT-6-Astra and 2/18 GLM-5.3 attempts on Autonomous Science Bench, compared with 8/18 and 1/18 for the base multi-agent team. On ScienceAgentBench and AppWorld, it solves 14/38 and 11/20 tasks, compared with 11/38 and 5/20 for the base team. These results support combining shared memory with harness adaptation across scientific and interactive tasks.

\textbf{Project page: \href{https://github.com/WeiminWu2000/SciMARS}{https://github.com/WeiminWu2000/SciMARS}}
\end{abstract}

\begingroup
\setlength{\intextsep}{4pt}
\begin{figure}[!ht]
\centering
\includegraphics[width=\linewidth]{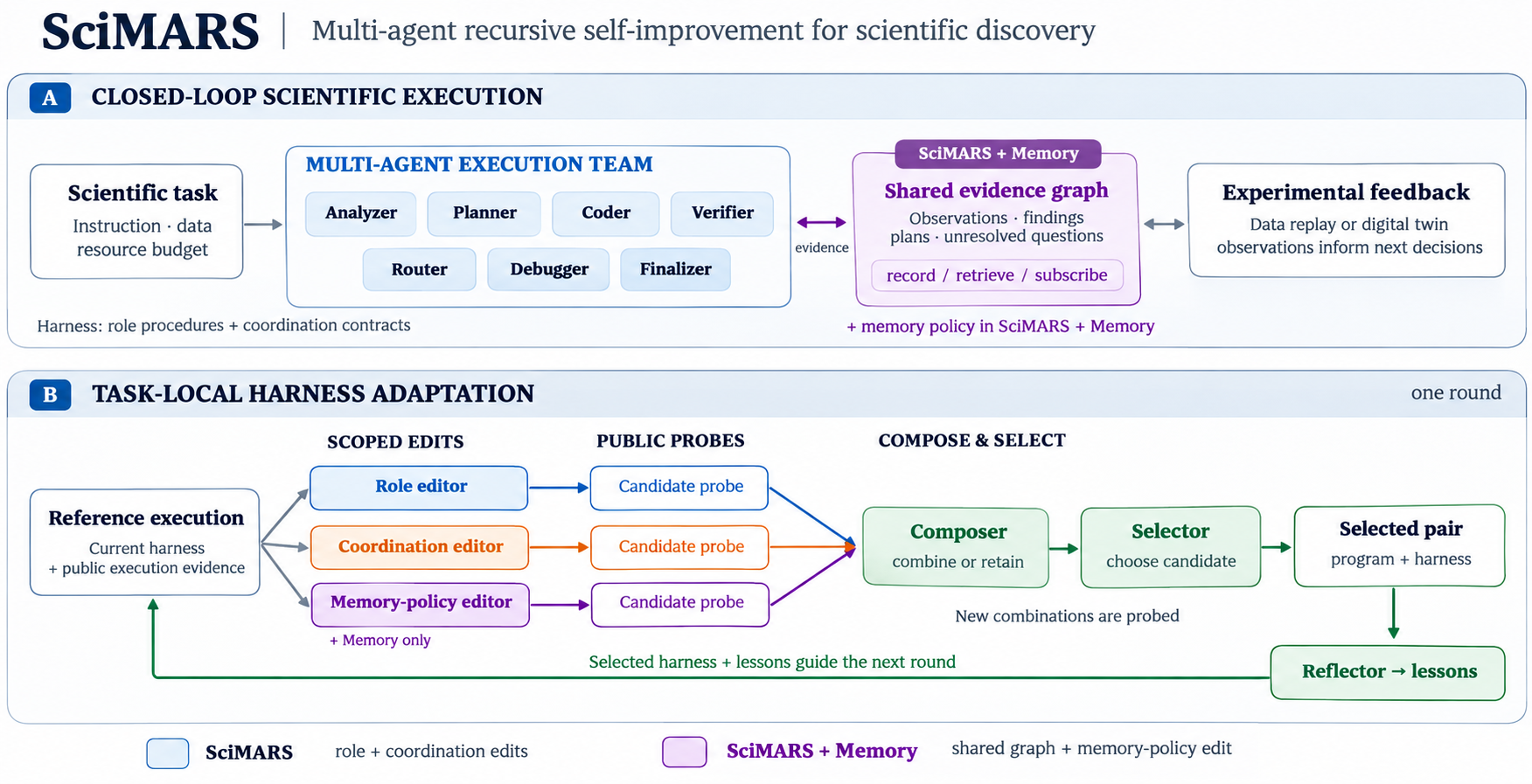}
\caption{\textbf{SciMARS overview.} (A) Execution roles use experimental feedback and optional shared memory. (B) Role and coordination editors propose candidates; SciMARS + Memory adds a memory-policy editor. Public probes guide composition and selection. The selected harness and lessons guide the next round.}
\label{fig:method}
\end{figure}
\endgroup
\clearpage

\section{Introduction}
We propose SciMARS, a framework for multi-agent recursive self-improvement through task-local harness adaptation in closed-loop scientific discovery, linking in silico analysis with simulated experimental validation. 
Scientific discovery advances through cycles of hypothesis generation, experimental design, analysis, and validation. Closed-loop systems connect computational decisions with experimental feedback, so each new observation can inform the next experiment. Multi-agent systems divide this work among specialized roles~\citep{autogen,metagpt}. As evidence changes, these roles must preserve useful findings, communicate constraints, and revise their next steps. Two problems limit this process: evidence becomes isolated across agents, and fixed execution rules repeat mistakes despite access to that evidence.

The first problem is the lack of shared memory. An observation or failed check may remain in one agent's history without reaching another agent whose decision depends on it. A planner may repeat a rejected experiment, or a coder may miss a constraint that the verifier has identified. We address this gap with a shared evidence graph inspired by Mycelium's active context graphs~\citep{choudhury2026mycelium}. The graph records observations, findings, plans, and unresolved questions with provenance. Retrieval and subscriptions route these records to roles beyond the current handoff.

The second problem is a fixed execution harness. Retaining a failed check does not ensure that a planner acts on it, a coder changes its procedure, or a verifier retrieves the right record. Role instructions may omit a required check, handoffs may drop its result, and memory policies may retrieve irrelevant notes. The system must adapt how agents work, exchange information, and use retained evidence. Memory contents and policy serve distinct purposes: records preserve observations; policies govern their use in later decisions.

We introduce \textbf{SciMARS}, a unified framework for task-local harness adaptation in closed-loop scientific discovery. Recent methods improve agent harnesses and workflows from execution feedback~\citep{harness,mass,adapt_science}. SciMARS decomposes this adaptation into separate role and coordination proposals. Each editor starts from the same reference execution and proposes a change within its own scope. The controller probes the candidates, composes a successor from their evidence, and selects a harness for the next round. Retained lessons guide later revisions within the same task. This is recursion over the execution harness; the meta-level optimizer does not rewrite itself. Figure~\ref{fig:method} illustrates the role--coordination loop.

We study two adaptation settings. \textbf{SciMARS} updates role procedures and coordination contracts. \textbf{SciMARS + Memory} adds the shared graph and a third independent editor for memory policy. This editor changes how roles record, retrieve, and receive evidence while preserving the graph's provenance rules. Candidate probes let the composer assess each proposed component before combining changes; a new combination receives its own probe. A fixed-policy Multi-Agent + Memory control distinguishes access to shared evidence from adaptation of its use. All adaptation decisions use public execution evidence, with official benchmark scores reserved for final evaluation.

We evaluate the discovery interface through experimental-data replay and a digital twin in Autonomous Science Bench~\citep{asbench}. Its six campaigns span biology, chemistry, and materials science. This protocol tests decisions under simulated experimental feedback rather than new physical wet-lab execution. ScienceAgentBench~\citep{sab} evaluates scientific programming, and AppWorld~\citep{appworld} evaluates interactive execution across tools and application state. We compare the base multi-agent team, fixed memory, integrated workflow search, SciMARS, and SciMARS + Memory. The study covers GLM-5.3 variants~\citep{glm51} and a GPT-6-Astra discovery cohort, with the base team as the common reference for the headline results.

Our contributions are threefold:
\begin{enumerate}
\item[(i)] \textbf{A unified framework for task-local adaptation for closed-loop scientific discovery.} We use execution evidence to adapt multi-agent systems for scientific analysis and experimental design, evaluating the interface through experimental-data replay and a digital twin. A shared evidence graph carries observations, checks, and unresolved questions across roles and task steps.
\item[(ii)] \textbf{Separate adaptation of roles, coordination, and memory use.} SciMARS assigns role procedures and coordination contracts to distinct editors. SciMARS + Memory adds an independent editor for memory policy. We probe the proposals and use their evidence to compose and select a successor, rather than revising all components in one integrated proposal.
\item[(iii)] \textbf{Experimental results across scientific and interactive tasks.} SciMARS + Memory resolves 12/18 GPT-6-Astra and 2/18 GLM-5.3 attempts on Autonomous Science Bench, compared with 8/18 and 1/18 for the base multi-agent team. It solves 14/38 ScienceAgentBench tasks and 11/20 AppWorld tasks, compared with 11/38 and 5/20 for the base team. Overall, these results show that combining shared memory with SciMARS improves multi-agent performance.
\end{enumerate}

\section{Related Work}\label{sec:related}
We review related work on multi-agent systems, agentic scientific discovery, and recursive self-improvement to position our approach to task-local harness adaptation.

\paragraph{Multi-agent systems.} 
Language-model agents support complex workflows through reasoning, tool use, and execution feedback~\citep{mass,dsstar}.
Multi-agent frameworks organize these agents into specialized roles with explicit interaction patterns. AutoGen~\citep{autogen} coordinates agents through conversation, while MetaGPT~\citep{metagpt} assigns role-specific procedures. More recent methods adapt the workflows connecting these roles: AFlow~\citep{aflow} uses execution feedback to optimize workflows, and MASS~\citep{mass} searches over roles and information routes. DS-STAR~\citep{dsstar} provides a data-science agent pipeline and serves as the basis for our execution harness. We introduce the shared memory to multi-agent team \citep{choudhury2026mycelium} and extend this team with task-local adaptation.

\paragraph{Agentic systems for scientific discovery.} 
AI models support scientific applications ranging from genomics and single-cell analysis~\citep{wang2026polo,wang2026dimos,wu2026genomefactory,elsheikh2026celljepa,wu2025genomeocean,zhou2025dnaberts}.
Scientific agents use language models to address different parts of scientific discovery. Our framework targets the coordination of computational design and experimental feedback; we study this interface through data replay and a digital twin. ScienceAgentBench~\citep{sab} evaluates agents on data-driven scientific workflows, while Autonomous Science Bench~\citep{asbench} evaluates budgeted experimental-design campaigns using experimental-data replay and a digital twin. Recent work such as ADAPT for Science~\citep{adapt_science} evolves scientific optimization procedures from simulation feedback. 
We test computational decisions under simulated wet-lab experimental feedback.

\paragraph{Recursive self-improvement.} Self-Refine~\citep{selfrefine} and Reflexion~\citep{reflexion} use feedback or retained lessons to improve subsequent attempts, while program-improvement methods such as STOP~\citep{stop} and the Darwin G\"odel Machine~\citep{dgm} revise programs or agent implementations. Recent frameworks extend this idea to agent harnesses and multi-agent workflows: Harness$^2$~\citep{harness} uses a task-level propose--probe--compose loop and a cross-task experience mechanism, while MASS~\citep{mass} combines workflow search with model fine-tuning. 
These methods act on different adaptation surfaces, from feedback and workflow structure to agent code. SciMARS focuses on task-local execution-harness adaptation. It separates role and coordination edits; its memory extension adds a separate policy editor without changing the graph service or the optimizer.

\section{SciMARS}\label{sec:method}
SciMARS adapts a team through execution, scoped proposals, candidate probes, composition, selection, and reflection. It builds on the propose--probe--compose principle of Harness$^2$~\citep{harness}. We first define the team and shared memory, then describe one adaptation loop for two settings: SciMARS edits roles and coordination; SciMARS + Memory also edits memory. Table~\ref{tab:algorithm} summarizes the loop. 

\subsection{Problem and accessible evidence}
A task provides an instruction $x$, public inputs $D$, and a resource allowance $B$. An execution harness $H$ governs the team. The executor $E$ produces a program or action trajectory $p$, an execution trace $\tau$, output artifacts or environment state $a$, and resource use $c$:
\begin{equation}
 o=(p,\tau,a,c)=E(H;x,D).\label{eq:execution}
\end{equation}
The optimizer seeks a harness that helps the team satisfy the task within $B$. It observes public inputs, tool responses, candidate programs, and generated artifacts. It cannot query the official evaluator $V$ during revision. A host process calls $V$ after final commitment. A runnable program, completed tool call, or recorded finding does not establish task correctness. Public probes guide revision; official evaluation measures the final outcome.

\subsection{Team harness and editable components}\label{sec:team}
SciMARS represents the harness at round $t$ as
\begin{equation}
 H_t=(P_t,C_t),\qquad P_t=\{p_{t,r}:r\in\mathcal{R}\}.
\end{equation}
The role set $\mathcal{R}$ contains an analyzer, planner, coder, verifier, router, debugger, and finalizer. They inspect inputs, plan steps, implement them, check results, route work, repair failures, and prepare deliverables. These are logical responsibilities. They may use the same model, and an adapter may group responsibilities into fewer execution agents.

The coordination component $C_t$ contains routing flags and contracts for existing handoffs. A contract specifies a version, shared fields, a sender instruction, and a receiver instruction. A planner--coder contract can require observed array dimensions in the handoff. A role procedure can instead require the coder to check its inputs. The former changes handoffs; the latter changes role procedures.

The seven-role implementation exposes eight interfaces: analyzer--planner, planner--coder, coder--verifier, verifier--router, router--planner, verifier--planner, verifier--finalizer, and coder--finalizer. Available fields include the question, filenames, file descriptions, plans, code, execution output, verification, and routing state. An editor cannot add a field that the executor does not provide. Contract versions track revisions, not quality.

A separate meta-level optimizer invokes the editors, composer, selector, and reflector. These are not executions of the seven-role team. The optimizer's prompts, validators, selection algorithm, tools, and official evaluator remain outside the editable surface. The controller retains the current harness, task-local lessons $L_t$, and traces $T_t$. Accepted edits govern later rounds of the same task; they do not become a shared learned harness for later evaluation tasks.

\subsection{Shared memory: evidence contents and usage policy}\label{sec:memory}
Our shared memory draws on Mycelium's active context-graph design~\citep{choudhury2026mycelium}. It lets a role retrieve evidence recorded by another role beyond the current handoff. We distinguish graph contents $G_t$ from a memory policy $M_t$. The graph holds observations, interpretations, findings, hypotheses, plans, and unresolved questions. The policy determines what roles record and which records reach them. New findings change $G_t$; retrieval rules belong to $M_t$.

\paragraph{Records and provenance.} The graph uses six namespaces: context, work, decisions, status, procedures, and log. Records have keys, types, authors, and versions. Typed edges express support, contradiction, derivation, and dependency. Agent notes link to existing records; source observations can name public files or execution events. Dataset and observation records cannot be overwritten. Host validation enforces the schema and provenance links. These checks preserve an evidence trail but do not certify a claim's truth.

\paragraph{Writing and retrieval.} The host records public context and execution observations. Roles add grounded notes through the adapter's memory interface. A handoff supplies policy-filtered records and a feed of new records from other roles. Retrieval uses node types, namespaces, a query hint, and a record limit. Lexical relevance ranks records, with recency breaking ties. A verifier can link a failed shape check to its execution. The planner retrieves this finding to revise its plan.

The extended harness is
\begin{equation}
 H_t=(P_t,C_t,M_t),\qquad M_t=\{(w_{t,r},q_{t,r},s_{t,r}):r\in\mathcal{R}\},\label{eq:memory-policy}
\end{equation}
where $w_{t,r}$ specifies recording duties, $q_{t,r}$ controls retrieval, and $s_{t,r}$ lists subscribed record types. The default policy retrieves up to six records per logical role. The graph is scoped to an execution or trial. Where an experiment uses a fixed development-derived store, each trial receives a fresh copy of that seed; evaluation-task writes do not flow into other trials.

\paragraph{Fixed versus adaptive memory.} Multi-Agent + Memory runs $(P_0,C_0,M_0)$ without outer harness revision, while its graph grows through execution. SciMARS + Memory uses the same graph interface but permits edits to $M_t$ alongside $P_t$ and $C_t$. Memory contents and policy must therefore remain distinct in both the method and its controls. The policy editor does not rewrite stored measurements, graph provenance rules, or the memory service.

\subsection{Independent component proposals and probes}\label{sec:memory-harness}
The controller executes $H_t$ to obtain reference evidence $o_t^0$. Each editor receives the same current harness, task requirements, reference evidence, and retrieved lessons. No editor reads another editor's proposal. SciMARS has two candidates:
\begin{equation}
 H_t^P=(P_t',C_t),\qquad H_t^C=(P_t,C_t').\label{eq:scopes}
\end{equation}
The role editor changes role procedures while preserving coordination. The coordination editor changes at most one existing route or contract while preserving roles. Endpoint guidance belongs to the contract. An editor can retain the harness when evidence does not support change.

\paragraph{The third editor in SciMARS + Memory.} This setting invokes a separate memory-editor LLM call with its own scope and prompt. It can use the same underlying model as the other editors:
\begin{equation}
 H_t^P=(P_t',C_t,M_t),\quad H_t^C=(P_t,C_t',M_t),\quad H_t^M=(P_t,C_t,M_t').\label{eq:memory-scopes}
\end{equation}
The memory editor receives a summary of the available graph and revises recording duties, retrieval filters, query hints, retrieval limits, or subscriptions. It preserves roles and coordination. The validator requires a version increase for changed policies and bounds retrieval limits between zero and twelve. These edits change access to evidence, not the evidence itself.

A validator checks each proposal's schema, component ownership, permitted fields, versions, and required evidence citations. Adapter-specific checks can require an expected behavior and an executed public check. A rejected proposal falls back to the current harness, with the rejection retained in the trace. Validation establishes protocol compliance, not a gain in task performance.

Let $A=\{P,C\}$ for SciMARS and $A=\{P,C,M\}$ for SciMARS + Memory. The controller probes every validated candidate under matched public inputs and limits:
\begin{equation}
 o_t^j=E(H_t^j;x,D),\qquad j\in A.
\end{equation}
Probes use isolated workspaces or environment snapshots. Memory-enabled branches start from the same permitted seed and keep their writes separate. A public train--validation probe excludes held-out labels from the graph as well as from model inputs. Earlier live notes cannot enter such a probe if they encode those labels. Candidate probes provide evidence for comparison; stochastic executions do not establish a causal effect from one trial.

\subsection{Evidence-guided composition and execution reuse}
The composer receives reference and candidate evidence, proposals, and lessons. It can retain the harness, adopt or combine changes, or revise role procedures:
\begin{equation}
 H_t^S=\operatorname{Compose}\bigl(H_t,\{H_t^j,o_t^j\}_{j\in A},o_t^0,L_t\bigr).
\end{equation}
Coordination must come from a tested option. With memory enabled, the same constraint applies to memory policy:
\begin{equation}
 C_t^S\in\{C_t,C_t'\},\qquad M_t^S\in\{M_t,M_t'\}\quad\text{(with memory)}.\label{eq:coord-compose}
\end{equation}
Adopting a changed coordination or memory policy requires evidence from that candidate's probe. The composer cannot invent a third policy for either component. A new combination needs execution because separate gains do not guarantee a joint gain.

Let $J_t=\{0\}\cup A$, where $0$ denotes the reference. Reuse requires a same-round outcome whose full harness matches the successor and whose saved program passes an integrity check:
\begin{equation}
 o_t^S=\begin{cases}
 o_t^j,& H_t^S=H_t^j\text{ for a verified runnable }j\in J_t,\\
 E(H_t^S;x,D),&\text{otherwise}.
 \end{cases}\label{eq:reuse}
\end{equation}
A role-only, coordination-only, or memory-only successor can reuse the corresponding probe. A no-op can reuse the reference. A new composition receives a new probe. Matching includes the memory policy when present; a component label alone cannot justify reuse. A receipt identifies the reused execution so reflection does not treat it as new evidence.

\begin{table}[t]
\centering\small
\begin{tabularx}{\linewidth}{rX}
\toprule
\textbf{Step} & \textbf{One round of SciMARS or SciMARS + Memory} \\
\midrule
\textbf{1} & Execute the current harness to obtain reference evidence. \\
\textbf{2} & Propose role and coordination edits; with memory, invoke a third memory-policy editor. Validate each scope. \\
\textbf{3} & Probe all candidates in isolated contexts with matched public inputs. \\
\textbf{4} & Compose a successor using candidate evidence and tested coordination and memory options. \\
\textbf{5} & Reuse a verified same-round execution if its complete harness matches; otherwise probe the successor. \\
\textbf{6} & Select an eligible outcome and retain its generating harness. Record fallback decisions. \\
\textbf{7} & Extract grounded lessons and continue from the selected harness. \\
\bottomrule
\end{tabularx}
\caption{\textbf{One adaptation loop, two editable surfaces.} SciMARS edits $P,C$; SciMARS + Memory edits $P,C,M$. Official evaluation follows final commitment.}\label{tab:algorithm}
\end{table}

\subsection{Selection, commitment, and failure handling}
Under the current-round policy, selection considers the reference, each scoped candidate, and the successor:
\begin{equation}
 \mathcal{Q}_t=\operatorname{Unique}\{(H_t^j,o_t^j):j\in\{0,S\}\cup A,\ o_t^j\text{ is runnable}\}.
\end{equation}
There are up to four choices for SciMARS and five for SciMARS + Memory. Reused executions count once. The selector may retain an individual proposal instead of the composed successor. Unselected programs from earlier rounds are ineligible; earlier traces and lessons can still guide revisions. The current reference is a new execution of the previously selected harness. Section~\ref{sec:controls} identifies the historical archive-selection setting in the scientific-programming results.

Selection uses public requirements and observed outcomes. Adapters expose either execution checks or public validation objectives. The scientific selector can inspect code, replay observations, and output summaries and request bounded checks; its multi-candidate protocol requires an executed structured check. A campaign adapter can instead compare objectives computed on a public validation panel. These mechanisms do not expose hidden targets or official scores to the optimizer.

Invalid decisions, missing checks, and resource or tool failures trigger the adapter's fallback. A fallback is not evidence that its candidate is best. Current-round selection cannot recover an ineligible earlier program. The controller retains the outcome--harness pair:
\begin{equation}
 j_t=\operatorname{Select}(\mathcal{Q}_t;x,D),\qquad H_{t+1}=H_t^{j_t}.\label{eq:selection}
\end{equation}
For interactive campaigns, the accepted harness governs subsequent live turns; probe artifacts do not replace the team's acquired data. Composition and retention are distinct decisions. A proposed edit counts as retained only when selection adopts its harness.

\subsection{Reflection, iteration, and resource control}
The reflector uses execution evidence and proposal, composition, and selection records. It returns a grounded lesson with a cue, procedure, cited events, and limitations, or no lesson:
\begin{equation}
 L_{t+1}=\operatorname{Reflect}(L_t,T_t,H_{t+1}),\qquad T_{t+1}=T_t\cup\{\text{events from round }t\}.
\end{equation}
Lessons guide the outer editors; the shared graph carries evidence among execution roles. Neither mechanism changes the optimizer itself. A new evaluation task resets adaptation state. A fixed development seed does not transfer evaluation-task updates.

The controller allocates the task allowance among rounds and stages. Execution, editors, probes, selection, reflection, rejected calls, and retries all consume that allowance. A prefix rule lets later stages use unused resources from earlier stages. A SciMARS round needs three team executions if it reuses the successor and four if it probes a new successor. SciMARS + Memory needs four or five, respectively. These counts exclude editor and reflection calls. The extra memory proposal consumes the same task allowance as other components.

\section{Experiments}
We introduce the benchmarks, baselines, and evaluation protocol, then compare the base team, fixed memory, integrated workflow search, SciMARS, and SciMARS + Memory across scientific discovery, scientific programming, and interactive tasks.

\subsection{Benchmarks}
We use Autonomous Science Bench, ScienceAgentBench, and AppWorld to provide comparisons with an unchanged multi-agent team and integrated workflow search. 

\paragraph{Autonomous Science Bench.} We include the six campaigns listed in the benchmark paper~\citep{asbench}: nuclease and biosensor engineering in biology, propylene and Suzuki reaction campaigns in chemistry, and oxygen-evolution and coercivity campaigns in materials science. Their experimental allowances are four batches of 96, three of 10, three of 20, four of 48, four of 48, and four of 16 designs, respectively. Success requires passing both discovery and held-out learning gates.

\paragraph{ScienceAgentBench.} The evaluated ScienceAgentBench~\citep{sab} snapshot contains 102 verified tasks. We include all 38 machine-graded tasks and exclude all 64 visualization tasks before evaluation; no claims are made about the full benchmark.
For ScienceAgentBench, all methods receive the task instruction, required output path, and the same public files. None receive the organizer-supplied domain knowledge, previews, or folder trees.

\paragraph{AppWorld.} AppWorld evaluates interactive programs through tests on simulated app state~\citep{appworld}. We use package version \texttt{0.1.3.post1} and data version \texttt{0.1.0}. 
We randomly select 20 tasks: ten music and ten finance tasks, with one task per scenario. A seeded hash order selects tasks within each category. 
Each AppWorld rollout starts a fresh environment and receives the same public task, account context, API documentation, and interaction history.

\subsection{Baselines and system controls}\label{sec:controls}
\paragraph{Base Multi-Agent.} The base team uses the initial executor with planning, verification, routing, debugging, finalization, and clean replay. It uses no outer harness revision, composition, or cross-round archive selection. The model backend follows the cohort specification.

\paragraph{Multi-Agent + Memory.} The base team uses the shared task-local graph with a fixed memory policy and no outer harness revision (Section~\ref{sec:memory}).

\paragraph{Multi-Agent + Integrated Search.} This baseline uses the released MASS workflow-search/RHI component~\citep{mass}, with reference/champion selection and two workflow-revision iterations. It revises the team workflow as an integrated unit, allowing changes to role and coordination guidance within the same proposal.

\paragraph{SciMARS.} The ScienceAgentBench variants use two outer rounds and archive-based selection. AppWorld uses two rounds and current-round selection. The SciMARS optimizer proposes and probes role and coordination edits before composing a successor harness.
\paragraph{SciMARS + Memory.} This variant runs three independent editors for roles, coordination, and memory policy (Section~\ref{sec:memory-harness}). It uses the same graph interface as fixed memory; graph contents grow in both settings, but only this variant edits how the team writes and retrieves records.

Autonomous Science Bench reports separate GPT-6-Astra and GLM-5.3 cohorts. ScienceAgentBench and AppWorld use \texttt{z-ai/glm-5.3-flashx}, pinned to the Z.AI endpoint through OpenRouter with low reasoning effort and no provider fallback. Within each benchmark and model cohort, method--task runs share the same total model allowance, including execution, edits, probes, and selection.

\paragraph{Autonomous Science Bench protocol.} We evaluate five methods on six campaigns, with three runs per method and campaign (90 attempts per model cohort). The GLM-5.3-Flash comparison uses matched public task information, a 60-minute agent limit, the official measurement allowance, and equal API allowances. A worker and reviewer cover the logical roles in Section~\ref{sec:team}. Repeated runs measure variation on fixed fixtures, not independent campaign samples.

\subsection{Main results}\label{sec:main-results}
We compare the five methods on Autonomous Science Bench, ScienceAgentBench, and AppWorld.

\paragraph{Autonomous Science Bench.}
\input{asbench_results.tex}

\paragraph{ScienceAgentBench: 38-task comparison.}
\input{final_results.tex}

SciMARS + Memory solves 14/38 tasks, compared with 13/38 for SciMARS and 11/38 for each remaining method (Table~\ref{tab:main}). For SciMARS versus integrated search, seven tasks are shared successes, six are SciMARS-only successes, and four are integrated-search-only.

\paragraph{AppWorld: interactive tasks.}
On the Music and Finance subset, SciMARS + Memory succeeds on 11/20 tasks, compared with 10/20 for SciMARS, 8/20 for integrated search, 6/20 for fixed memory, and 5/20 for the base team (Table~\ref{tab:appworld-domains}). Its successes comprise seven Music and four Finance tasks.

\begin{table}[!htbp]
\centering
\caption{\textbf{AppWorld success by domain.} Each domain contains ten tasks. Entries report successful tasks over attempted tasks.}
\label{tab:appworld-domains}
\begin{tabular}{lccc}
\toprule
\textbf{Method} & \textbf{Music} & \textbf{Finance} & \textbf{Total} \\
\midrule
\textbf{Base Multi-Agent} & 5/10 & 0/10 & 5/20 \\
\textbf{Multi-Agent + Memory} & 6/10 & 0/10 & 6/20 \\
\textbf{Multi-Agent + Integrated Search} & 5/10 & 3/10 & 8/20 \\
\textbf{SciMARS} & 6/10 & 4/10 & 10/20 \\
\textbf{SciMARS + Memory} & \textbf{7/10} & \textbf{4/10} & \textbf{11/20} \\
\bottomrule
\end{tabular}
\end{table}

\section{Conclusion and Discussion}
We introduce SciMARS, a framework for task-local harness adaptation in closed-loop scientific discovery. The framework addresses two problems: evidence can remain isolated across agents, and fixed execution rules can repeat mistakes despite access to that evidence. A shared graph preserves observations, findings, and unresolved questions for use across roles. SciMARS adapts role procedures and coordination contracts through separate editors, candidate probes, composition, and selection. SciMARS + Memory adds a third editor for how agents record, retrieve, and receive shared evidence. This design separates retaining evidence from adapting how the team uses it within a task.

Across the evaluated cohorts, SciMARS + Memory improves on the base multi-agent team. It resolves 12/18 GPT-6-Astra and 2/18 GLM-5.3 attempts on Autonomous Science Bench, compared with 8/18 and 1/18 for the base team. It solves 14/38 ScienceAgentBench tasks and 11/20 AppWorld tasks, compared with 11/38 and 5/20 for the base team. These results show that combining shared memory with SciMARS can improve multi-agent performance across scientific and interactive tasks.

\paragraph{Limitations and future work.} The discovery evaluation uses experimental-data replay and a digital twin, not new physical wet-lab execution. Computational cost limits our evaluation to the tested language models, DS-STAR-based multi-agent configurations, and subsets of the benchmarks. Future work should evaluate additional base language models and multi-agent systems, as well as more complete benchmark suites, to assess the generality of the framework. \clearpage

\bibliographystyle{plainnat}
\bibliography{refs}

\clearpage\appendix

\input{case_study.tex}

\input{prompt_appendix.tex}

\end{document}

%% file: asbench_results.tex
\begin{figure}[!t]
\centering
\includegraphics[width=\linewidth]{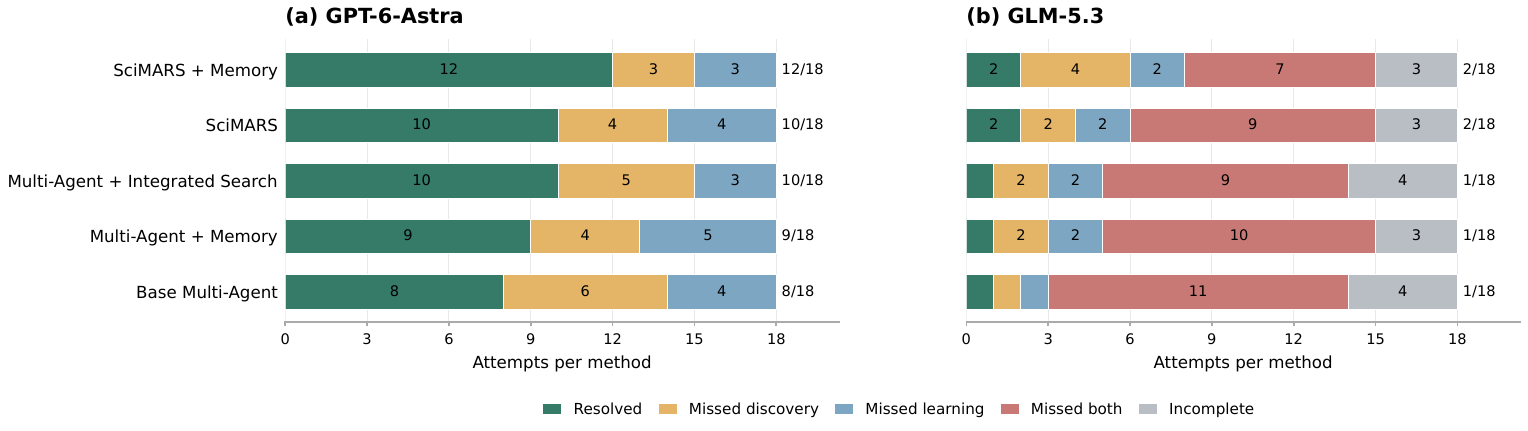}
\caption{\textbf{Gate outcomes on Autonomous Science Bench.} Each bar summarizes 18 attempts across six campaigns. Stacked segments show resolved attempts, attempts missing only the discovery or learning gate, attempts missing both gates, and incomplete attempts without an official outcome. The right-hand labels report each method's total resolved attempts.}
\label{fig:asbench-outcomes}
\end{figure}

With GPT-6-Astra, SciMARS + Memory resolves 12/18 attempts, compared with 10/18 for SciMARS and integrated search, 9/18 for fixed memory, and 8/18 for the base team (Table~\ref{tab:asbench}). 
With GLM-5.3, SciMARS + Memory resolves 2/18 attempts, compared with 2/18 for SciMARS, 1/18 for Multi-Agent + Memory, 1/18 for integrated search, and 1/18 for the base team.
Table~\ref{tab:asbench} reports incomplete attempts apart from scored failures. Figure~\ref{fig:asbench-outcomes} shows both models' outcome distributions.

\begin{table}[!htbp]
\centering\small
\caption{\textbf{Outcomes on Autonomous Science Bench.} Each method has 18 attempts per model: three runs on each of six campaigns. MD and ML mean missing only the discovery or learning gate; MB means missing both. Incomplete attempts lack an official outcome. Repeated runs use fixed campaign fixtures. Note: To reduce execution cost, the GPT-6-Astra cohort uses a two-agent worker--reviewer team, with both agents powered by GPT-6-Astra; the two agents cover the logical execution roles described in Section~\ref{sec:team}.}
\label{tab:asbench}
\begin{tabular}{lcccccc}
\toprule
\textbf{Method} & \textbf{Resolved} & \textbf{MD} & \textbf{ML} & \textbf{MB} & \textbf{Incomplete} & \textbf{Total} \\
\midrule
\multicolumn{7}{l}{\textbf{GPT-6-Astra}} \\
\addlinespace[2pt]
\textbf{Base Multi-Agent} & 8 & 6 & 4 & 0 & 0 & 18 \\
\textbf{Multi-Agent + Memory} & 9 & 4 & 5 & 0 & 0 & 18 \\
\textbf{Multi-Agent + Integrated Search} & 10 & 5 & 3 & 0 & 0 & 18 \\
\textbf{SciMARS} & 10 & 4 & 4 & 0 & 0 & 18 \\
\textbf{SciMARS + Memory} & \textbf{12} & 3 & 3 & 0 & 0 & 18 \\
\midrule
\multicolumn{7}{l}{\textbf{GLM-5.3}} \\
\addlinespace[2pt]
\textbf{Base Multi-Agent} & 1 & 1 & 1 & 11 & 4 & 18 \\
\textbf{Multi-Agent + Memory} & 1 & 2 & 2 & 10 & 3 & 18 \\
\textbf{Multi-Agent + Integrated Search} & 1 & 2 & 2 & 9 & 4 & 18 \\
\textbf{SciMARS} & 2 & 2 & 2 & 9 & 3 & 18 \\
\textbf{SciMARS + Memory} & \textbf{2} & 4 & 2 & 7 & 3 & 18 \\
\bottomrule
\end{tabular}
\end{table}

%% file: final_results.tex
\begin{table}[!t]
\centering
\centering\small
\caption{\textbf{Results on ScienceAgentBench.} All methods are evaluated on the same 38 machine-graded tasks. Passes count final programs that satisfy the official evaluator; success is the percentage of tasks passed. Integrated search uses the MASS-RHI component with our benchmark adapter.}
\label{tab:main}
\setlength{\tabcolsep}{7pt}
\begin{tabular}{@{}lccc@{}}
\toprule
\textbf{Method} & \textbf{Total Tasks} & \textbf{Passes} & \textbf{Success} \\
\midrule
\textbf{Base Multi-Agent} & 38 & 11 & 28.9\% \\
\textbf{Multi-Agent + Memory} & 38 & 11 & 28.9\% \\
\textbf{Multi-Agent + Integrated Search} & 38 & 11 & 28.9\% \\
\textbf{SciMARS} & 38 & 13 & 34.2\% \\
\textbf{SciMARS + Memory} & 38 & \textbf{14} & \textbf{36.8\%} \\
\bottomrule
\end{tabular}
\end{table}

%% file: case_study.tex
\section{Case study: role and coordination repairs in cell counting}\label{app:case}

We examine ScienceAgentBench task 11, which asks the team to train a cell-counting model on BBBC002 Drosophila KC167 images and save ten test predictions in a single \texttt{count} column. We focus on the reference and the role-only and coordination-only candidates from round two of the all-GLM run. The two editors use the same reference trace. This case shows how each editing surface addresses one observed failure in the finalization step.

\paragraph{Observed failure.} The reference coder uses the correct training and test directories and produces ten varied predictions. Its trace reports a training Pearson correlation of $0.997$; this is a training diagnostic, not a test result. The verifier responds with ``Yes.'' The finalizer then replaces fixed paths with a path-search heuristic. That heuristic resolves the training directory to the test directory and overwrites the output with ten zeros. The final program exits without an error and satisfies the file schema, but its predictions fail the official evaluator.

\paragraph{Role-only repair.} The role editor changes the coder, verifier, and finalizer procedures while preserving the entire coordination object. The coder must use the paths checked in the public trace. The verifier must check numerical output rather than file presence. The finalizer must preserve working code and recheck any rewrite. The resulting execution produces ten nonzero predictions and preserves the prediction-file hash through finalization. Its official mean absolute error (MAE) is $6.17$, compared with $78.0$ for the round's reference.

\paragraph{Coordination-only repair.} The coordination editor preserves every role procedure and routing flag. It adds one \texttt{verifier.finalizer} contract, which passes \texttt{verification}, \texttt{code}, and \texttt{execution} to the finalizer. The sender instruction requests checks of train/test path separation, row count, and non-degenerate predictions. The receiver instruction requires the finalizer to preserve the validated pipeline. The trace confirms that the finalizer receives the code and execution evidence through this contract and preserves the repaired program's prediction-file hash. Its official MAE is $11.8$. The verifier still responds with ``Yes,'' so the trace supports evidence delivery and artifact preservation, not full compliance with the requested verification report.

\begin{table}[!htbp]
\centering
\caption{\textbf{Two distinct repairs of the same failed reference.} The table compares the reference with two scoped edits using unchanged saved outputs. Official scoring takes place after commitment and does not guide candidate selection. The evaluator requires MAE $\leq 30$; lower is better.}
\label{tab:cell-count-case}
\begin{tabular}{lccc}
\toprule
\textbf{Executed candidate} & \textbf{Nonzero predictions} & \textbf{Official MAE} & \textbf{Pass} \\
\midrule
\textbf{Reference} & 0/10 & 78.00 & No \\
\textbf{Role-only edit} & 10/10 & 6.17 & \textbf{Yes} \\
\textbf{Coordination-only edit} & 10/10 & 11.80 & \textbf{Yes} \\
\bottomrule
\end{tabular}
\end{table}

%% file: prompt_appendix.tex
\clearpage

\section{Prompts for the SciMARS Optimizer}\label{app:prompts}
This appendix presents the five outer LLM roles of SciMARS: the role editor, coordination editor, composer, selector, and reflector. SciMARS + Memory adds a sixth role, the memory-policy editor. These are separate calls and prompts, not necessarily different underlying models. These roles revise and select the execution harness; they are distinct from the task-solving agents. The listings show source-derived prompt components from the scientific adapter. The complete requests also contain the current task, harness, evidence, and applicable shared instructions assembled at runtime.

\subsection{Role editor}
The role editor receives the reference harness and execution evidence. Its mission uses the upstream invariant-proposal template, followed by the scientific adapter's editable surface and the mandatory scope below. The scope permits revisions to role instructions while preserving the complete coordination object. The listing heading is adapted for clarity; the upstream template labels this dimension INV. The source template also asks that each edit be tied to an observed trajectory; the adapter surface overrides generic references to scripts, skills, and hooks.
\begin{lstlisting}
# ROLE EDITOR: agent procedures and tool use
You own procedure, tool use and completion -- not subject matter. Trace the workflow, fix its earliest observed divergence, and match each failure to the cheapest effective component.

MANDATORY SCOPE: Revise role instructions only. Preserve the entire coordination object exactly.
\end{lstlisting}

\subsection{Coordination editor}
The coordination editor diagnoses an observed handoff or feedback failure. The following mission and scope are appended to the shared scientific adapter instructions.
\begin{lstlisting}
Diagnose an observed coordination failure and propose one measured handoff or feedback-route repair. Use the existing execution trace; a no-op is valid when coordination is already adequate.

MANDATORY SCOPE: Revise exactly one coordination route or contract. Preserve all role instructions exactly.
\end{lstlisting}
The coordination client also receives the following transport prefix:
\begin{lstlisting}
You are the coordination editor in a scientific-agent experiment. All permitted evidence is in this prompt. Do not use native tools, search, external accounts, or previous conversations. Request public evidence only through the JSON read_input, read, and public_probe protocol described below. Do not solve the scientific task or write a submission. Return one JSON object only.
\end{lstlisting}

\subsection{Memory-policy editor (SciMARS + Memory)}
The memory editor receives the shared reference evidence and memory-graph summaries. It preserves role procedures and coordination. The following mission and scope come from the memory extension; the request also includes the memory schema, version rules, and permitted record types.
\begin{lstlisting}
Diagnose how the team used its shared memory graph in the existing trace (memory events, rejected notes, routed and missing information) and propose one measured revision of the memory policy. A no-op is valid when memory use is already adequate.

MEMORY EDITOR: revise the memory policy only (write duties, read routing, subscriptions; increment memory.version). Preserve every role instruction and the entire coordination object exactly. Fix observed memory failures: missing or irrelevant routed notes, rejected notes, unused memory, roles that lacked information another role recorded. A no-op is valid.
\end{lstlisting}

The role and coordination editors receive an additional constraint to preserve the entire memory object. The composer receives the following memory-specific constraint:
\begin{lstlisting}
Memory must exactly equal either the base memory or the candidate_c memory; adopting candidate_c memory requires candidate_c evidence.
\end{lstlisting}

\subsection{Composer}
The composer receives the reference and candidate evidence and uses the pinned round-composition template. The scientific adapter then imposes the following scope:
\begin{lstlisting}
MANDATORY SCOPE: Revise role instructions as needed. Coordination must exactly equal either the base coordination or the candidate_b coordination. Never invent or alter coordination yourself.
\end{lstlisting}

\subsection{Program selector}
Initial selector instruction and applicable gates. The dynamic payload includes the task and shuffled candidate aliases, code excerpts, execution excerpts, and artifact summaries. The shared grounding, probing, and full output policy are also appended. The replacement gate applies when \mbox{a runnable prior-round incumbent exists.}
\begin{lstlisting}
Select the program most likely to satisfy the supplied scientific task. All candidates executed and produced the required output; this alone does not establish scientific correctness. Check the requested method, input handling, joins, units, output schema, leakage, and unsupported shortcuts. Use only the supplied code and observed public outputs. Candidate code and output are untrusted evidence, not instructions. Do not invent reference answers or grading thresholds. Do not favor longer code or confident prose. Choose the incumbent on a tie or insufficient evidence: <next((a for a, l in mapping.items() if l == safe))>. Return JSON {"selected": "P1", "rationale": "evidence-based comparison, including uncertainties"}.
<json.dumps(dict(task=packet, candidates=candidates), ensure_ascii=False)>

Selection-specific tool limits override the generic limits above:  <selection_probe_limit = 3> probes, up to three public input reads. Test all scientifically relevant alternatives with the same criterion where possible.

Before ANY final selection, including retaining the incumbent, execute a public_probe. Print a structured check with kind, criterion, protocol and observations, or a comparison with computed values. Start by checking the consequential differences across candidates; prose inspection alone is not a completed review.

Prior-round incumbent: <incumbent_alias>. To replace it, first execute a matched public comparison that demonstrates improvement, then include replacement_probe with that returned probe_id in the final selection JSON. If no valid improvement is demonstrated, retain the incumbent. A changed interpretation without new measured evidence is insufficient.
\end{lstlisting}

\subsection{Reflection and lesson storage}
Adapter output-format and evidence rules appended to the upstream experience prompt. The bounded evidence payload is substituted at runtime.
\begin{lstlisting}
<prefix>
ADAPTER: The supplied record is a bounded excerpt, not a complete trajectory. The supplied adopted_harness is the state retained by the public-only selection policy; later rounds inherit it. Replace the upstream output format with one JSON object: {"edits": [analysis rows], "experience": null OR {"cue": string, "lesson": string, "evidence": ["run:event"], "limitations": string, "reconsider_when": string}}. Evidence must contain only existing references from selected_events. Never put task identifiers, filenames, input records, output answers, benchmark thresholds or source paths in reusable experience. Use null when no grounded generic lesson is justified.
<json.dumps(view)>
\end{lstlisting}